# A Bayesian Approach toward Active Learning for Collaborative Filtering

**Rong Jin**

Department of Computer Science and Engineering
Michigan State University
rong@cse.cmu.edu

**Luo Si**

Language Technology Institute
School of Computer Science
Carnegie Mellon University
Pittsburgh, PA 15232
lsi@cs.cmu.edu

**Abstract**

Collaborative filtering is a useful technique for exploiting the preference patterns of a group of users to predict the utility of items for the active user. In general, the performance of collaborative filtering depends on the number of rated examples given by the active user. The more the number of rated examples given by the active user, the more accurate the predicted ratings will be. Active learning provides an effective way to acquire the most informative rated examples from active users. Previous work on active learning for collaborative filtering only considers the expected loss function based on the estimated model, which can be misleading when the estimated model is inaccurate. This paper takes one step further by taking into account of the posterior distribution of the estimated model, which results in more robust active learning algorithm. Empirical studies with datasets of movie ratings show that when the number of ratings from the active user is restricted to be small, active learning methods only based on the estimated model don't perform well while the active learning method using the model distribution achieves substantially better performance.

## 1. Introduction

The rapid growth of the information on the Internet demands intelligent information agent that can sift through all the available information and find out the most valuable to us. Collaborative filtering exploits the preference patterns of a group of users to predict the utility of items for an active user. Compared to content-based filtering approaches, collaborative filtering systems have advantages in the environments where the contents of items are not available due to either a privacy issue or the fact that contents are difficult for a computer to analyze (e.g. music and videos). One of the key issues in the collaborative filtering is to identify the group of users who share the similar interests as the active user. Usually, the similarity between users are measured based on their ratings over the same set of items. Therefore, to accurately identify users that share similar interests as the active user, a reasonably large number of ratings from the active user are usually required. However, few users are willing to provide ratings for a large amount of items. Active learning methods provide a solution to this problem by acquiring the ratings from an active user that are most useful in determining his/her interests. Instead of randomly selecting an item for soliciting the rating from the active user, for most active learning methods, items are selected to maximize the expected reduction in the predefined loss function. The commonly used loss functions include the entropy of model distribution and the prediction error. In the paper by Yu et. al. (2003), the expected reduction in the entropy of the model distribution is used to select the most informative item for the active user. Boutilier et. al. (2003) applies the metric of expected value of utility to find the most informative item for soliciting the rating, which is essentially to find the item that leads to the most significant change in the highest expected ratings.

One problem with the previous work on the active learning for collaborative filtering is that computation of expected loss is based only on the estimated model. This can be dangerous when the number of rated examples given by the active user is small and as a result the estimated model is usually far from being accurate. A better strategy for active learning is to take into account of the model uncertainty by averaging the expected loss function over the posterior distribution of models. With the full Bayesian treatment, we will be able to avoid the problem caused by the large variance in the model distribution. Many studies have been done on the active learning to take into account of the model uncertainty. The method of query by committee (Seung et. al., 1992; Freud et. al., 1996) simulates the posterior distribution of models by constructing an ensemble of models and the example with the largest uncertainty in prediction is selected for user's feedback. In the work by Tong and Koller (2000), a full Bayesian analysis of the active learning for parameter estimation in Bayesian Networks is used, which takes into account of the model uncertainty in computing the loss function. In this paper, we will apply the full Bayesian analysis to the active learning for collaborative filtering. Particularly, in order to simplify the computation, we approximate the posterior distribution of model with a simple Dirichlet distribution, which leads to an analytic expression for the expected loss function.

The rest of the paper is arranged as follows: Section 2 describes the related work in both collaborative filtering and active learning. Section 3 discusses the proposed active learning algorithm for collaborative filtering. The experiments are explained and discussed in Section 4. Section 5 concludes this work and the future work.

## 2. Related Work

In this section, we will first briefly discuss the previous work on collaborative filtering, followed by the previous work on active filtering. The previous work on active learning for collaborative filtering will be discussed at the end of this section.

### 2.1 Previous Work on Collaborative Filtering

Most collaborative filtering methods fall into two categories: Memory-based algorithms and Model-based algorithms (Breese et al. 1998). Memory-based algorithms store rating examples of users in a training database. In the predicating phase, they predict the ratings of an active user based on the corresponding ratings of the users in the training database that are similar to the active user. In contrast, model-based algorithms construct models that well explain the rating examples from the training database and apply the estimated model to predict the ratings for active users. Both types of approaches have been shown to be effective for collaborative filtering. In this subsection, we focus on the model-based collaborative filtering approaches, including the Aspect Model (AM), the Personality Diagnosis (PD) and the Flexible Mixture Model (FMM).

For the convenience of discussion, we will first introduce the annotation. Let items denoted by $X = \{x_1, x_2, ......, x_M\}$, users denoted by $Y = \{y_1, y_2, ......, y_N\}$, and the range of ratings denoted by $\{1,...,R\}$. A tuple $(x, y, r)$ means that rating $r$ is assigned to item $x$ by user $y$. Let $X(y)$ denote the set of items rated by user $y$, and $R_y(x)$ stand for and the rating of item $x$ by user y, respectively.

**Aspect model** is a probabilistic latent space model, which models individual preferences as a convex combination of preference factors (Hofmann & Puzicha 1999; Hofmann, 2003). The latent class variable $z \in Z = \{z_1, z_2, ......, z_K\}$ is associated with each pair of a user and an item. The aspect model assumes that users and items are independent from each other given the latent class variable. Thus, the probability for each observation tuple $(x, y, r)$ is calculated as follows:

$$p(r \mid x, y) = \sum_{z \in Z} p(r \mid z, x) p(z \mid y) \quad (1)$$

where $p(z|y)$ stands for the likelihood for user $y$ to be in class $z$ and $p(r|z,x)$ stands for the likelihood of assigning item $x$ with rating $r$ by users in class z. In order to achieve better performance, the ratings of each user are normalized to be a normal distribution with zero mean and variance as 1 (Hofmann, 2003). The parameter $p(r|z,x)$ is approximated as a Gaussian distribution $N(\mu_z, \sigma_z)$ and $p(z|y)$ as a multinomial distribution.

**Personality diagnosis approach** treats each user in the training database as an individual model. To predicate the rating of the active user on certain items, we first compute the likelihood for the active user to be in the 'model' of each training use, which is approximated using a Gaussian distribution:

$$p(y' \mid y) \propto \prod_{x \in X(y')} e^{-(R_y(x) - R_{y'}(x))^2 / 2\sigma^2} \quad (2)$$

where $\sigma$ stands for the variance of Gaussian distributions. The ratings from the training user on the same items are then weighted by the computed likelihood. The weighted average is used as the estimation of ratings for the active user. Previous empirical studies have shown that the personality diagnosis method is able to outperform several other approaches for collaborative filtering (Pennock et al., 2000).

**Flexible Mixture Model** introduces two sets of hidden variables $\{z_x, z_y\}$, with $z_x$ for the class of items and $z_y$ for the class of users (Si and Jin, 2003; Jin et. al., 2003). Similar to aspect model, the probability for each observed tuple $(x, y, r)$ is factorized into a sum over different classes for items and users, i.e.,

$$p(x, y, r) = \sum_{z_x, z_y} p(z_x) P(z_y) P(x \mid z_x) P(y \mid z_y) P(r \mid z_x, z_y) \quad (3)$$

All the parameters are estimated using Expectation Maximization algorithm (EM) (Dumpster et. al., 1976). The multiple-cause vector quantization (MCVQ) model (Boutilier and Zemel, 2003) uses the similar idea for collaborative filtering.

### 2.2 Previous Work on Active Learning

The goal of active learning is to learn the correct model using only a small number of labeled examples. The general approach is to find the example from the pool of unlabeled data that gives the largest reduction in the expected loss function. The loss functions used by most active learning methods can be categorized into two groups: the loss functions based on model uncertainty and the loss function based on prediction errors. For the first type of loss functions, the goal is to achieve the largest reduction ratio in the space of hypothesis. One commonly used loss function is the entropy of the model distribution. Methods within this category usually select the example for which the model has the largest uncertainty in predicting the label (Seung et. al., 1992; Freud et. al, 1997; Abe and Mamitsuka, 1998; Campbell et. al., 2000; Tong and Koller, 2002). The second type of loss functions involved the prediction errors. The largest reduction in the volume of hypothesis space may not necessary be effective in cutting the prediction error. Freud et. al. showed an example in (1997), in which the largest reduction in the space of hypothesis does not bring the optimal improvement to the reduction of prediction errors. Empirical studies in Bayesian Network (Tong and Koller, 2000) and text categorization (Roy and MaCallum, 2001) have shown that using the loss function that directly targets on the prediction error is able to achieve better performance than the loss function that is only based on the model uncertainty. The commonly used loss function within this category is the entropy of the distribution of predicted labels.

In addition to the choice of loss function, how to estimate the expected loss is another important issue for active learning. Many active learning methods compute the expected loss only based on the currently estimated model without taking into account of the model uncertainty. Even though this simple strategy works fine for many applications, it can be very misleading, particularly when the estimated model is far from the true model. As an example, considering learning a classification model for the data distribution in Figure 1, where spheres represent data points of one class and stars represent data points of the other class. The four labeled examples are highlighted by the line-shaded ellipsis. Based on these four training examples, the most likely decision boundary is the horizontal line (i.e., the dash line) while the true decision boundary is a vertical line (i.e., the dot line). If we only rely on the estimated model for estimating the expected loss, the examples that will be selected for user's feedback are most likely from the dot-shaded areas, which are ineffective in adjusting the estimated decision boundary (i.e. the horizontal line) to the correct decision boundary (i.e. the vertical). On the other hand, if we can use the model distribution for computing the expected loss, we will be able to adjust the decision boundary more effectively since decision boundaries other than the estimated one (i.e. horizontal line) are considered in the computation of expected loss.

There have been several studies on active learning that utilize the posterior distribution of models for estimating the expected loss. The query by committee approach simulates the model distribution by sampling a set of models out of the posterior distribution. In the work of active learning for parameter estimation in Bayesian Network, a Dirichlet distribution for the parameters is used to estimate the change in the entropy function. However, the general difficulty with the full Bayesian analysis for active learning is the computational complexity. For complicated models, usually it is rather difficult to obtain the exact posterior distribution for models. As a result, sampling approaches such as Markov Chain Mote Carlo (MCMC) and Gibbs sampling are used to approximate the Bayesian average. In this paper, we follow an idea similar to the Laplace approximation used in graphic model (MacKay, 1992). Instead of accurately computing the posteriors, we approximate the posterior distribution with an analytic expression and apply the approximated posterior distribution to estimate the expected loss. Compared to the sampling approaches, this approach substantially simplifies the computation by avoiding generating a large number of models and calculating the loss function values over the generated models.

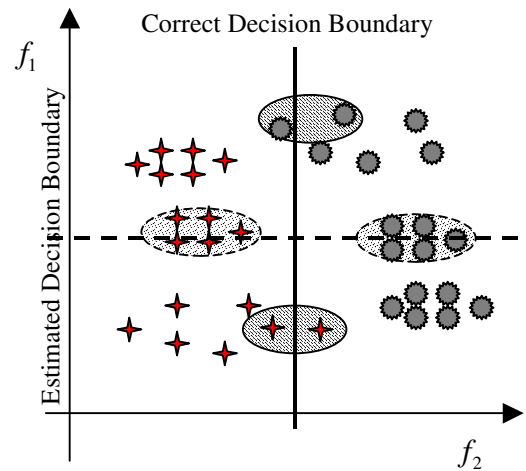

**Figure 1**: A learning scenario when the estimated model is far from the true model. The vertical line corresponds to the correct decision boundary and the horizontal line corresponds to the estimated decision boundary. The four labeled examples are highlighted by the line-shaded areas.

### 2.3 Previous Work on Active Learning for Collaborative Filtering

There have been only a few studies on active learning for collaborative filtering. In the paper by Kai et al (2003), a method similar to Personality Diagnosis (PD) is used for collaborative filtering. Each example is selected for user's feedback in order to reduce the entropy of the like-mindness distribution or $p(y|y')$ in Equation (2). In the paper by Boutilier et al (2003), the Multiple-cause vector quantification method (similar to the FMM model) is used for collaborative filtering. Unlike many other collaborative filtering researches, which try to minimize the prediction error, this paper only concerns with the items that are strongly recommended by the system. The loss function is based on the expected value of information (EVOI), which is computed based on the currently estimated model. One problem with both studies on active learning for collaborative filtering is that, the expected loss is computed only based on a single model, namely the currently estimated model. As illustrated by the example in Figure 1, the estimation based on only a single model can be misleading, particularly when the number of rated examples given by the active user is small and meanwhile the number of parameters to be estimated is large. In the late experiments, we will show that the selection based on the expected loss function using only the currently estimated model can be even worse than simple random selection.

## 3. A Bayesian Approach Toward Active Learning for Collaborative Filtering

In this section, we will first discuss the approximated analytic expression for the posterior distribution. Then, the approximated posterior distribution will be used to estimate the expected loss.

### 3.1 Approximate the Model Posterior Distribution

For the sake of simplicity, we will focus on the active learning of aspect model for collaborative filtering. However, the principle used in this section can be easily extended to other models for collaborative filtering.

As described in Equation (1), each conditional likelihood $p(r|x,y)$ is decomposed as the sum of $\sum_z p(r|z,x)p(z|y)$. For active user $y'$, the most important task is to determine its user type, or $p(z|y')$. Let $\theta = \{\theta_z = p(z|x)\}_z$ stands for the parameter space. Usually, parameters $\theta$ are estimated through the maximum likelihood estimation, i.e.

$$\theta^* = \arg\max_{\theta \in \Theta} \prod_{x \in X(y')} p(r|x,y';\theta)$$
$$= \arg\max_{\theta \in \Theta} \prod_{x \in X(y')} \sum_{z \in Z} p(r|z,x)\theta_z \quad (4)$$

Then, the posterior distribution of model, i.e., $p(\theta|D(y'))$ where $D(y')$ includes all the ratings given the active user $y'$, can be written as:

$$p(\theta|D(y')) = \frac{1}{Z(D(y'))} p(D(y')|\theta)p(\theta)$$
$$= \frac{1}{Z(D(y'))} \prod_{x \in X(y')} \sum_{z \in Z} p(r|z,x)\theta_z \quad (5)$$

where $Z(D(y'))$ is the normalization factor. In above, a uniform distribution is used for prior $p(\theta)$. Apparently, the posterior distribution in (7) is rather difficult to be used for estimating the expected loss due to the multiple products and the normalization factor $Z(D(y'))$.

To approximate the posterior distribution, we will consider the expansion of the posterior function around the maximum point. Let $\theta^*$ stand for the maximal parameters that are obtained from EM algorithm by maximizing Equation (4). Let's consider the ratio of $\log p(\theta|D(y'))$ with respect to $\log p(\theta^*|D(y'))$, which can be written as:

$$\log \frac{p(\theta|D(y'))}{p(\theta^*|D(y'))} = \sum_i \log \frac{\sum_z p(r_i|x_i,z)\theta_z}{\sum_z p(r_i|x_i,z)\theta_z^*}$$
$$\approx \sum_i \sum_z \frac{p(r_i|x_i,z)\theta_z^*}{\sum_{z'} p(r_i|x_i,z')\theta_{z'}^*} \log \frac{\theta_z}{\theta_z^*} \quad (6)$$
$$= \sum_z (\alpha_z - 1) \log \frac{\theta_z}{\theta_z^*}$$

where $\alpha_z$ is defined as

$$\alpha_z = \sum_i \frac{p(r_i|x_i,z)\theta_z^*}{\sum_{z'} p(r_i|x_i,z')\theta_{z'}^*} + 1 \quad (7)$$

Based on the approximation in Equation (8), the approximated posterior distribution $p(\theta|D(y'))$ is a Dirichlet with hyper parameters $\alpha_z$ defined in Equation (9), or

$$p(\theta|D) \approx \frac{\Gamma(\alpha^*)}{\prod_z \Gamma(\alpha_z)} \prod_z \theta^{\alpha_z - 1} \quad (8)$$

where $\alpha^* = \sum_z \alpha_z$. Furthermore, the following relation between the hyper parameters $\alpha_z$ and optimal parameters $\theta^*$ is true:

$$\frac{\alpha_z - 1}{\alpha_{z'} - 1} = \frac{\theta_z^*}{\theta_{z'}^*} \quad (9)$$

This is because the optimal parameters obtained by the EM algorithm is actual the fixed point of the following equation:

$$\theta_z^* = \sum_i \frac{p(r_i|x_i,z)\theta_z^*}{\sum_{z'} p(r_i|x_i,z')\theta_{z'}^*} \quad (10)$$

Based on the property in Equation (9), it is not difficult to verify that the optimal point for the approximated Dirichlet distribution in Equation (8) is exactly $\theta^*$.

### 3.2 Estimate the Expected Loss

Similar to many other active learning works (Tong and Koller, 2000; Seung et al, 1992; Kai et al, 2003), we use the entropy of the model function as the loss function. Many previous studies on active learning try to find the example that directly minimizes the entropy of the model parameters with currently estimated model. In the case of aspect model, it can be formulated as the following optimization problem:

$$x^* = \arg\min_{x \in X} - \left\langle \sum_z \theta_{z|x,r} \log \theta_{z|x,r} \right\rangle_{p(r|x,\theta)} \quad (11)$$

where $\theta$ denotes the currently estimated parameters and $\theta_{z|x,r}$ denotes the optimal parameter that are estimated using one additional rated example, i.e., item $x$ is rated as $r$. The goal of Equation (11) is to find item $x$ such that the expected entropy of distribution $\theta$ is minimized. There are two problems with applying this simple strategy to collaborative filtering:

1) The first problem comes from the fact that the expected entropy is computed only based on the currently estimated model. According to Equation (11), the expectation is evaluated over the distribution $p(r|x,\theta)$. As already pointed out in Section 2, such estimation could be misleading particularly when the currently estimated model is far from the true model.

2) The second problem comes from the characteristics of collaborative filtering. Recall that parameter $\theta_z$ stands for the likelihood for the active user to be in the user class $z$. Since the optimization in Equation (11) tries to efficiently reduce the uncertainty in the type of user class for the active user, it will select examples that can "purify" the class distribution $\theta_z$. This contradicts the previous studies in collaborative filtering (Si and Jin, 2003), which found it is more effective for collaborative filtering to assume a user of multiple user types than a single type. As a result of this observation, we would expect to the distribution $\theta$ be of multiple modes, not of a single model.

Based on the above analysis, the optimization problem in Equation (11) is not appropriate for active learning of collaborative filtering. The later empirical studies also indicate this fact.

In the ideal case, if the true model $\theta^{true}$ is given, the optimal strategy in selecting example should be:

$$x^* = \arg\min_{x \in X} - \left\langle \sum_z \theta_z^{true} \log \frac{\theta_{z|x,r}}{\theta_z^{true}} \right\rangle_{p(r|x,\theta^{true})} \quad (12)$$

The goal of Equation (12) is to find an example such that the updated model parameter $\theta_{z|x,r}$ can be adjusted toward the true model parameter $\theta_z^{true}$ most efficiently. Since the true model $\theta^{true}$ is unknown, we need to approximate the optimization goal in Equation (12). One way of approximation is to replace the true model $\theta^{true}$ with the expectation over the posterior distribution $p(\theta | D(y'))$. As a result of this approximation, Equation (12) is transformed into the following optimization problem:

$$x^* = \arg\min_{x \in X} - \left\langle \left\langle \sum_z \theta_z' \log \frac{\theta_{z|x,r}}{\theta_z'} \right\rangle_{p(r|x,\theta')} \right\rangle_{p(\theta|D(y))} \quad (13)$$

Now, the key issue becomes how to compute the integration efficiently since the sampling methods are usually time consuming. Fortunately, the integration in Equation (13) can be computed analytically as follows:

$$\left\langle \left\langle \sum_z \theta_z' \log \theta_{z|x,r} \right\rangle_{p(r|x,\theta')} \right\rangle_{p(\theta|D(y))}$$

$$= \sum_r \int p(\theta' | D(y)) p(r|x,\theta') \sum_j \theta_j' \log \frac{\theta_{j|r,x}}{\theta_j'} d\theta'$$

$$= \sum_r \sum_i \sum_j \left\langle p(r|x,i) \theta_i' \theta_j' \log \frac{\theta_{j|r,x}}{\theta_j'} \right\rangle_{p(\theta|D(y))}$$

$$= \frac{\Psi(\alpha^* + 2)}{\alpha^*} \sum_r \sum_i \alpha_i p(r|x,i)$$

$$- \frac{\sum_j \alpha_j \Psi(\alpha_j + 1)}{(\alpha^* + 1)\alpha^*} \sum_r \sum_i \alpha_i p(r|x,i) \quad (14)$$

$$- \frac{\sum_r \sum_j \alpha_j \Psi(\alpha_j + 1) p(r|x,j)}{(\alpha^* + 1)\alpha^*}$$

$$+ \sum_r \frac{\sum_j \alpha_j \log \theta_{j|x,r} \sum_i \alpha_i p(r|x,i)}{(\alpha^* + 1)\alpha^*}$$

$$+ \sum_r \frac{\sum_j \alpha_j \log \theta_{j|x,r} p(r|x,j)}{(\alpha^* + 1)\alpha^*} - \frac{\sum_r \sum_i \alpha_i p(r|x,i)}{(\alpha^* + 1)\alpha^*}$$

where $\Psi$ is the digamma function. The above derivation uses the following property of digamma function:

$$\left\langle \log \theta_j \right\rangle_{\theta \sim Dirichlet(\vec{\alpha})} = \Psi(\alpha_j) - \Psi(\alpha^*)$$

Note that $\sum_{r=1}^{R} p(r|x,\theta) \neq 1$ since a Gaussian approximation is used to compute $p(r|x,\theta)$. With the analytic result in Equation (14), the integration in Equation (13) can be efficiently computed.

The other computational complexity comes from the estimation of the updated model $\theta_{z|x,r}$. The standard way to obtain the updated model is to rerun the full EM algorithm with one more additional rated example (i.e., example $x$ is rated as category $r$). This can be extremely expensive since we will have an updated model for every item and every possible rating category. A more efficient way for computing the updated model is to avoid the directly optimization of Equation (4). Instead, we use the following approximation,

$$\begin{aligned}\theta^* &= \arg\max_{\theta \in \Theta} p(r | x, \theta) \prod_{x' \in X(y')} p(r | x', y'; \theta) \\ &= \arg\max_{\theta \in \Theta} p(r | x, \theta) p(\theta | D(y')) \\ &\approx \arg\max_{\theta \in \Theta} p(r | x, \theta) Dirichlet(\vec{\alpha}) \\ &\approx \arg\max_{\theta \in \Theta} \sum_{z'} p(r | x, z') \theta_{z'} \prod_{z} \theta_z^{\alpha_z - 1}\end{aligned} \quad (15)$$

The EM updating equation for the above objective function is:

$$\theta_z = \frac{p(r_i | x_i, z)\theta_z + \alpha_z - 1}{\alpha^* + \sum_{z'}(p(r_i | x_i, z')\theta_{z'} - 1)} \quad (16)$$

The advantage of the updating equation in (16) versus the more general EM updating equation in (10) is that it only depends on the rating $r$ and item $x$ while Equation (10) has to go through all the items rated by the active user.

In summary, we propose a Bayesian treatment of active learning for collaborative filtering, which uses the model posterior distribution to compute the estimation of loss function. To simplify the computation, we approximate the mode distribution with a Dirichlet distribution. As a result, the expected loss can be calculated analytically and the user model can be updated more efficiently. For later reference, we call this method '**Bayesian method**'.

## 4. Experiments

In this section, we present experiment results in order to address the following two questions:

1) *Whether the proposed active learning algorithm is effective for collaborative filtering?* In the experiment, we will compare the proposed algorithm to the method of randomly acquiring examples from the active user.

2) *How important is the full Bayesian treatment?* In this paper, we emphasize the importance of taking into account the model uncertainty using the posterior distribution. To illustrate this point, in this experiment, we compare the proposed algorithm to two commonly used active learning methods that are only based on the estimated model without utilizing the model distribution.

**Table 1**: Characteristics of MovieRating and EachMovie.

|  | MovieRating | EachMovie |
|---|---|---|
| Number of Users | 500 | 2000 |
| Number of Items | 1000 | 1682 |
| Avg. # of rated Items/User | 87.7 | 129.6 |
| Number of Ratings | 5 | 6 |

The details of these two active learning methods will be discussed later.

### 4.1 Experiment Design

Two datasets of movie ratings are used in our experiments, i.e., 'MovieRating'[1] and 'EachMovie'[2]. For 'EachMovie', we extracted a subset of 2,000 users with more than 40 ratings. The details of these two datasets are listed in Table 1. For 'MovieRating' dataset, we use the first 200 users for training and the rest users for testing. For 'EachMovie' dataset, the first 400 users are used for training. For each test user, we randomly selected three items with their ratings as the starting seed for the active learning algorithm to build the initial model. Furthermore, for each test user, twenty rated items are reserved for evaluating the performance of different active learning methods. Finally, for each iteration, an active learning algorithm is allowed to solicit rating for a single item from the active user. Totally, it can ask for ratings of five different items and the performance is evaluated for each feedback. For simplicity, we assume that the active user will always be able to rate the items presented by the active learning. Of course, as pointed out in (Kai et al, 2003), this is not a completely correct assumption because there are items that the active user has not seen before and therefore it is impossible for him/her to rate those items. Since the focus of this paper is on the behavior of active learning algorithms for collaborative filtering, we will leave this issue for future work.

The aspect model used for collaborative filtering has already been described in the Section 2. The number of user classes used in the experiment is set to be 5 for MovieRating dataset and 10 for EachMovie datasets based on the previous empirical studies.

The mean absolute error (MAE) is used to evaluate the performance of collaborative filtering, which is defined as follows (Breese et al., 1998):

$$MAE = \frac{1}{L_{Test}} \sum_{l} | r_{(l)} - \hat{R}_{y_{(l)}}(x_{(l)}) | \quad (17)$$

where $L_{Test}$ is the number of the test ratings.

---

[1] http://www.cs.usyd.edu.au/~irena/movie_data.zip

[2] http://research.compaq.com/SRC/eachmovie

The proposed algorithm is compared against the following three active learning algorithms for collaborative filtering:

1) *Random Selection*: This method randomly selects one item out of the pool of items for user's feedback. We refer this simple method as '**random method**'.

2) *Model Entropy based Sample Selection*. This method has already been described at the beginning of Section 3.2. It finds the item that efficiently reduces the entropy of the user class distribution for the active user (in Equation (11)). As aforementioned, the problems with this simple selection strategy are in two folds: I) Computing the expected loss only based on the currently estimated model and, II) Conflicting with the intuition that each user can be of multiple types. By comparing this approach to the proposed active learning method for collaborative filtering, we will be able to see if the idea of using model distribution for computing expected loss is worthwhile for collaborative filtering. We will refer to this method as '**model entropy method**'.

3) *Prediction Entropy based Sample Selection*. Unlike the previous method, which concerns with the uncertainty in assigning the active user to different user classes, this method focuses on the uncertainty in predicting ratings of items for the active user. It selects the item that is able to reduce the entropy of predicating ratings for different items. Formally, the selection criterion can be formulated as the following optimization problem:

$$x^* = \arg\min_{x \in X} - \left\langle \sum_{x'} p(r|x', \theta_{|x,r}) \log p(r|x', \theta_{|x,r}) \right\rangle_{p(r|x,\theta)} \quad (18)$$

where notation $\theta_{|r,x}$ stands for the updated model using the extra rated example $(r,x,y)$. Similar to the previous method, this approach uses the estimated model for computing the expected loss. However, unlike the previous approach that tries to 'purify' the distribution of user types for the active user, this approach targets on the prediction distribution. Therefore, it is *not* against the intuition that each user is of multiple types. We will refer this method as '**prediction entropy method**'.

**4.2 Results and Discussion**

The results for the proposed active learning algorithm together with the three baseline models for database 'MovieRating' and 'EachMovie' are presented in Figure 2 and 3, respectively.

First, according to Figure 2 and 3, both the 'model entropy method' and the 'prediction entropy method' performs consistently worse than the simple 'random method'. This phenomenon can be explained by the fact that the initial number of rated items given by the active user is only three, which is relatively small given the number of parameters to determine is 5 for 'MovieRating' and 10 for 'EachMovie'. As a result, the expected loss can not be computed accurately based on the estimated model, and the selected items will not be the most informative ones. Furthermore, comparing to the 'prediction entropy method', we see that the 'model entropy method' performs substantially worse. For example, in Figure 3, for the 'model entropy method', the performance of collaborative filtering is almost unchanged while the 'predication entropy method' is able to reduce the MAE error from 1.15 to 1.03. This is because the 'model entropy method' tries to narrow down a single user type for the active user. Since most users are of multiple types. it is inappropriate to apply the 'model entropy method' to collaborative filtering. On the other hand, the 'prediction entropy method' doesn't have this defect because it focuses on minimizing the uncertainty in predicating ratings of items for the active user instead of the uncertainty in assigning user types to the active user.

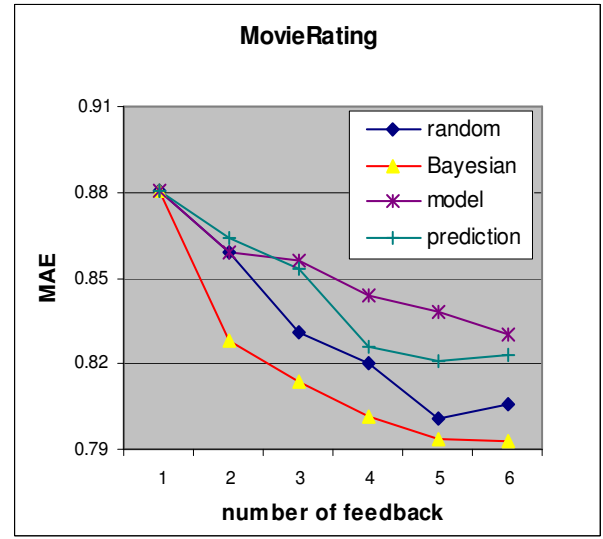

**Figure 2**: MAE results of four active learning algorithms for collaborative filtering over 'MovieRating' dataset. Legend 'random' stands for the random method, 'Bayesian' for the Bayesian method, 'model' for the model entropy method, and 'predication' for the prediction entropy method. The smaller the MAE the better the performance.

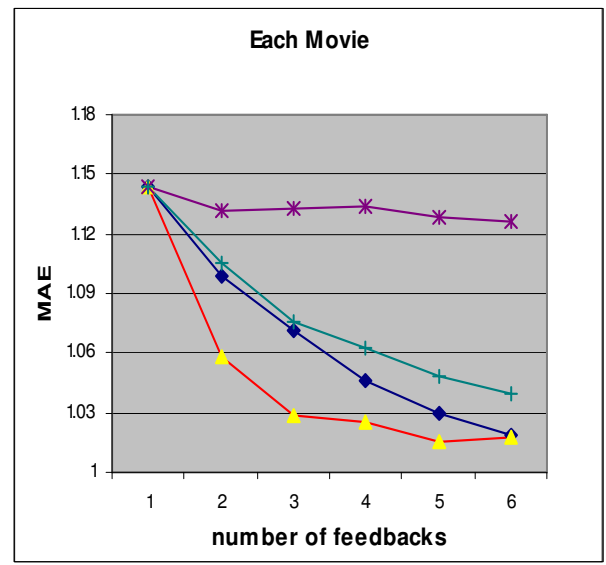

**Figure 3**: MAE results of four active learning algorithms for collaborative filtering over 'EachMovie' dataset. Legend 'random' stands for the random method, 'Bayesian' for the Bayesian method, 'model' for the model entropy method, and 'predication' for the prediction entropy method. The smaller the MAE the better the performance.

The second observation from Figure 2 and 3 is that the proposed active learning method performs better than any of the three based line models for both 'MovieRating' and 'EachMovie' datasets. The most important difference between the proposed method and the other methods for active learning is that the proposed method takes into account the model distribution, which makes it robust when even there is only three rated items given by the active user.

## 5. Conclusion

In this paper, we proposed a full Bayesian treatment of active learning for collaborative filtering. Different from previous studies of active learning for collaborative filtering, this method takes into account the model distribution when computing the expected loss. In order to alleviate the computation complexity, we approximate the model posterior with a simple Dirichlet distribution. As a result, the estimated loss can be computed analytically and the model for the active user can be updated efficiently. Since this work only focuses on the model quality, we plan to apply the full Bayesian analysis to the prediction error, which usually is more effective according to the previous studies of active learning.